\documentclass[runningheads]{llncs}

\usepackage{eccv}

\usepackage{eccvabbrv}

\usepackage{url}
\usepackage{xcolor}         
\usepackage{booktabs}    
\usepackage{multirow}    
\usepackage{graphicx}    
\usepackage{amsmath}     
\usepackage{wrapfig}
\usepackage{float}
\usepackage{algorithm}
\usepackage{algorithmicx}
\usepackage{algpseudocode}
\usepackage{amssymb}
\usepackage{array} 
\usepackage{adjustbox}  
\usepackage[table]{xcolor}
\usepackage{colortbl}
\usepackage{pgfplots}
\usepackage{makecell}
\pgfplotsset{compat=1.18}
\usepgfplotslibrary{groupplots}

\usepackage[accsupp]{axessibility}  

\usepackage{hyperref}
\usepackage{orcidlink}

\begin{document}

\title{\textit{Salt:} Self-Consistent Distribution Matching with Cache-Aware Training for Fast Video Generation}

\titlerunning{Salt}

\author{Xingtong Ge\inst{1,2}\orcidlink{0009-0007-8114-870X} \and
Yi Zhang\inst{2}\orcidlink{0000-0001-7147-9125} \and
Yushi Huang\inst{1}\orcidlink{0009-0002-7898-8402} \and 
Dailan He\inst{2}\orcidlink{0009-0002-6712-0616} \and
Xiahong Wang\inst{2} \and
Bingqi Ma\inst{2} \and
Guanglu Song\inst{2} \and
Yu Liu\inst{2} \and
Jun Zhang\inst{1}\orcidlink{0000-0002-5222-1898}\thanks{Corresponding Author}} 
\authorrunning{Xingtong Ge et al.}

\institute{Hong Kong University of Science and Technology \and
Vivix Group Limited \\
\email{xingtong.ge@gmail.com, eejzhang@ust.hk} \\
}
\maketitle

\begin{abstract}
Distilling video generation models to extremely low inference budgets (e.g., 2--4 NFEs) is crucial for real-time deployment, yet remains challenging. Trajectory-style consistency distillation often becomes conservative under complex video dynamics, yielding an over-smoothed appearance and weak motion. Distribution matching distillation (DMD) can recover sharp, mode-seeking samples, but its local training signals do not explicitly regularize how denoising updates compose across timesteps, making composed rollouts prone to drift. 
To overcome this challenge, we propose Self-Consistent Distribution Matching Distillation (SC-DMD), 
which explicitly regularizes the endpoint-consistent composition of consecutive denoising updates. For real-time autoregressive video generation, we further treat the KV cache as a quality parameterized condition and propose Cache-Distribution-Aware training. This training scheme applies SC-DMD over multi-step rollouts and introduces a cache-conditioned feature alignment objective that steers low-quality outputs toward high-quality references. Across extensive experiments on both non-autoregressive backbones (e.g., Wan~2.1) and autoregressive real-time paradigms (e.g., Self Forcing), our method, dubbed \textbf{Salt}, consistently improves low-NFE video generation quality while remaining compatible with diverse KV-cache memory mechanisms.
Project page: \href{https://xingtongge.github.io/Salt}{https://xingtongge.github.io/Salt}.

\end{abstract}

\section{Introduction}
\label{sec:intro}

Despite rapid progress in diffusion and flow-matching models~\cite{ho2020denoising,lipmanflow,liu2022flow,wan2025wan}, pushing video generation to extremely low inference budgets (e.g., 2--4 NFEs) remains challenging. A natural strategy, inherited from few-step image generation, is to distill a multi-step teacher into a few-step student using trajectory-level consistency objectives~\cite{song2023consistency,luo2023latent,geng2025mean,frans2024one,cai2025shortcutting}. However, video dynamics are inherently more complex, as the same conditioning may admit multiple plausible futures---for example, the continuation of a motion across time can vary in speed, extent, or trajectory. Under regression-style trajectory losses, this ambiguity tends to pull the predictor toward conditional averages, often resulting in over-smoothed appearance and conservative motion~\cite{hastie2009elements,zhai2024motion,mao2025osv}.

By contrast, distribution-level score distillation---notably Distribution Matching Distillation (DMD)~\cite{wang2023prolificdreamer,yin2024one,yin2024improved}---has emerged as a widely adopted paradigm for practical video distillation. By matching the student's output distribution to a strong reference through adversarial or distribution-matching gradients, DMD avoids regressing toward a single target trajectory and often produces sharper, more expressive samples in the low-NFE regime. This advantage has driven rapid adoption across recent video distillation methods~\cite{cheng2025phased,nie2026transition,lightx2v,lin2025diffusion,huang2025linvideo,lin2025autoregressive,huang2026qvgenpushinglimitquantized}. It has also become a core ingredient in autoregressive real-time generation systems---including CausVid~\cite{yin2025slow}, Self Forcing~\cite{huang2025self}, and LongLive~\cite{yang2025longlive}---suggesting that DMD-style objectives have become a practical default for few-step video generation. Crucially, most of these systems operate at 2--8 denoising steps rather than a single step, meaning that inference quality depends on repeatedly composing the learned denoising operator across multiple steps.


However, this reliance on multi-step composition exposes a structural gap in the DMD framework.
Few-step sampling is fundamentally a problem of repeatedly composing learned denoising updates, which is actually incompatible with the objective of predicting clean outputs in DMD training. As a result, per-step errors can accumulate along the composed trajectory, leading to drift as the number of steps increases~\cite{sabouralign}. In this work, we note that the training objective for DMD is purely local: each noise level is supervised independently as a ``one-step generator'', with no constraint on how the per-timestep behaviors interact when applied in sequence.
We term this \emph{compositionality deficit}: despite strong single-step quality, increasing the number of denoising steps can degrade rather than improve output (e.g., over-exposure and loss of fine-grained semantics; see Fig.~\ref{fig:dmd-error}).
This issue may be further amplified in autoregressive settings~\cite{huang2025self,yang2025longlive,zhu2026causal}, where each newly generated chunk is conditioned on previous generations; errors introduced in earlier chunks can be encoded into the generated KV cache and progressively propagated to later chunks over time.


In this work, we propose \textbf{Self-Consistent Distribution Matching Distillation (SC-DMD)}. The key idea is to retain DMD as the primary driver of distribution alignment and single-step sharpness, while augmenting it with a \emph{shortcut self-consistency regularizer} to mitigate the compositionality deficit. Concretely, this regularizer penalizes the semigroup defect of the student-induced transport operator, encouraging endpoint consistency under the composition of consecutive student updates. Rather than replacing distribution matching with a trajectory-style objective, we use this constraint as a complementary structural bias: the self-consistency term improves compositionality, while DMD preserves sharpness and avoids the mode-averaging tendencies of pure trajectory-based supervision in video. This combination yields consistent improvements in both non-autoregressive video generation (e.g., Wan~2.1~\cite{wan2025wan}) and autoregressive real-time generation.
Moreover, autoregressive real-time video generation introduces an additional challenge beyond few-step sampling. With KV caching, each newly generated chunk is conditioned on previous generations, and the quality of this conditioning depends on how accurately earlier chunks were denoised; consequently, errors can propagate through the cache and accumulate over long rollouts~\cite{huang2025self,yin2025slow}. We find that training with a single fixed step count leaves the generator brittle to the cache-quality variation encountered at inference. To address this, we introduce \textbf{Cache-Aware Mixed-Step Training} for autoregressive generation. Specifically, training alternates among rollouts with $K\in\{2,4,8\}$ denoising steps, exposing the model to cache-conditioned inputs of varying quality and significantly improving few-step generation with the application of self-consistency mechanism. Besides, this mixed-step setup naturally enables cross-step supervision: under the same noise and context, higher-step outputs can be used to regularize lower-step outputs through \emph{cache-conditioned feature alignment}. By transferring richer information from high-step predictions to low-step ones, this design further improves extreme low-step performance and rollout stability.
In summary, our contributions are threefold:
\begin{itemize}
    \item We identify the \emph{compositionality deficit} of DMD: although distribution-level score distillation is widely used, its purely local training signal imposes no structural constraint on multi-step coherence.
    \item We propose SC-DMD, which addresses the compositionality deficit by introducing a semigroup-defect regularizer. It improves multi-step stability while preserving the sharpness of distribution matching, yielding consistent gains in both non-autoregressive and autoregressive video generation.
    \item We further propose a cache-aware mixed-step training strategy for autoregressive distillation, combining mixed-step rollout with self-consistency and cache-conditioned representation alignment. This further improves extreme low-step generation and long-horizon rollout stability.
\end{itemize}

Across extensive experiments on both non-autoregressive (e.g., Wan~2.1) and autoregressive real-time paradigms (e.g., Self Forcing and LongLive), our full method, dubbed \textbf{Salt}, consistently improves extremely low-NFE video generation quality and long-horizon stability in autoregressive rollouts, while retaining compatibility with diverse KV-cache memory mechanisms~\cite{huang2025self,yang2025longlive,lu2025reward} and incurring no additional inference overhead.

\section{Related Work}

\noindent\textbf{Diffusion Distillation}
methods mainly fall into two categories: trajectory-based~\cite{song2023consistency,luo2023latent,wang2024phased,Lv_2025_ICCV,sabouralign,liu2025see} and distribution-based approaches~\cite{yin2024one,yin2024improved,ge2025senseflow,wang2025vdot,lin2025diffusion,cheng2025phased,nie2026transition,huang2025harmonicaharmonizingtraininginference,lightx2v,he2025neighbor}. Trajectory-based distillation compresses sampling by approximating the teacher’s denoising dynamics: DCM~\cite{Lv_2025_ICCV} mitigates the detail degradation by decoupling semantic learning from detail refinement via a dual-expert design. rCM~\cite{zheng2025large} scales continuous-time consistency distillation to application-scale image and video diffusion through efficient JVP training, and incorporates score-distillation as a long-skip regularizer to preserve fine details under few steps generation. 
Distribution-based methods instead target matching the generative distribution: Recent one-step video methods~\cite{lin2025diffusion,lin2025autoregressive} operationalize distribution matching via adversarial post-training, while POSE~\cite{cheng2025phased} introduces a phased equilibrium procedure to stabilize adversarial one-step distillation for video models.

\noindent\textbf{Autoregressive Video Generation}
enables streaming and interactivity via causal factorization, and can be grouped into training-based~\cite{teng2025magi,chen2025skyreels,yin2025slow,huang2025self,liu2025rolling,yang2025longlive,lu2025reward} and training-free~\cite{yesiltepe2025infinity,zhao2025ultravico} approaches. Training-based methods close the train–test gap of long rollouts beyond standard teacher forcing: Self Forcing~\cite{huang2025self} directly trains on autoregressive self-rollouts so each step conditions on previously generated context, reducing exposure bias; Causal Forcing~\cite{zhu2026causal} further pinpoints the bidirectional-to-causal distillation gap and uses an autoregressive teacher for ODE initialization before asymmetric DMD~\cite{yin2024one}, leading to higher-quality real-time interactive generation; Reward Forcing~\cite{lu2025reward} distills bidirectional video diffusion into a few step autoregressive student with rewarded distribution matching to improve efficient streaming generation. In addition, long-horizon AR generation emphasize scalable causal designs and long-video interaction mechanisms: MAGI-1~\cite{teng2025magi} scales chunk-wise autoregressive denoising for strong temporal consistency and deployability, LongLive~\cite{yang2025longlive} targets real-time interactive long videos with mechanisms for stable prompt transitions and long-range consistency under causal decoding.
On the other hand, training-free methods, such as Infinity-RoPE~\cite{yesiltepe2025infinity} modifies temporal encoding and KV-cache behaviors to unlock action-controllable infinite self-rollout, while UltraViCo~\cite{zhao2025ultravico} studies length extrapolation failure modes and suppresses out-of-window attention to reduce repetition and quality degradation.


\section{Method}

\subsection{Preliminaries}

\noindent\textbf{Diffusion/Flow-Matching Models and Flow Maps.}
Let $x_t \in \mathbb{R}^d$ denote a noisy latent at noise level $t\in[0,1]$ (larger $t$ indicates noisier states), and $c$ denote conditioning (e.g., text, image, or autoregressive context).
Many diffusion and flow-matching models admit a \emph{probability flow ODE} view, where sampling corresponds to transporting states from high noise to low noise via a learned vector field:
\begin{equation}
\frac{d x_t}{d t} = v(x_t, t, c), \qquad t\in[0,1], \quad x_{t_0}\sim p(x_{t_0}).
\label{eq:pf_ode}
\end{equation}
The induced timestep-to-timestep transport operator (flow map) deterministically maps a state at timestep $t$ to timestep $s<t$:
\begin{equation}
x_s = \Phi^{t\rightarrow s}(x_t;\,c), \qquad 0\le s < t \le 1.
\label{eq:flow_map_def}
\end{equation}
A fundamental property of exact ODE flow maps is the \emph{composition (semigroup) law}:
\begin{equation}
\Phi^{t\rightarrow s} = \Phi^{u\rightarrow s}\circ \Phi^{t\rightarrow u}, \qquad \forall~ t>u>s,
\label{eq:semigroup}
\end{equation}
which has been explicitly exploited by recent flow-map learning and distillation frameworks~\cite{boffi2024flow,boffibuild,sabouralign,kimctm}.
In practice, $\Phi$ is approximated by a numerical solver (e.g., Euler) using a learned vector field $v_\theta$.

\noindent\textbf{Distribution Matching Distillation (DMD).}
Assume a pre-trained teacher diffusion (or flow-matching) model is available, inducing a reference
distribution $p_r(x_t)$ and a teacher score $s_r(x_t,t)=\nabla_{x_t}\log p_r(x_t)$.
Distribution Matching Distillation (DMD) distills a few-step generator by matching the student
distribution to the teacher distribution at sampled noise levels.
Instead of directly optimizing $D_{\mathrm{KL}}(p_g\|p_r)$, DMD introduces an intermediate
``fake'' distribution $p_f(x_t,t)$ parameterized by a learnable score
$s_\phi(x_t,t)=\nabla_{x_t}\log p_f(x_t,t)$, and alternates between updating the critic ($\phi$)
and the generator ($\theta$).

In practice, the generator produces a clean sample $x_0 = G_\theta(z)$ with $z\!\sim\!\mathcal{N}(0,I)$, and a noisy sample at level $t$ is obtained by forward diffusion $x_t \sim q(x_t\mid x_0)$ (e.g., $x_t=\alpha(t)x_0+\sigma(t)\epsilon$, $\epsilon\!\sim\!\mathcal{N}(0,I)$), where $x_t$ depends on $\theta$ through $x_0$.
DMD yields a generator update direction of the form
\begin{equation}
\nabla_{\theta}\mathcal{L}_{\mathrm{DMD}}
~=~
\mathbb{E}_{t,z,\epsilon}\Big[
\big(s_{\phi}(x_t,t)-s_{r}(x_t,t)\big)\,
\frac{\partial x_t}{\partial \theta}
\Big],
\label{eq:dmd_grad_prelim}
\end{equation}


\subsection{SC-DMD: Self-Consistent Distribution Matching Distillation}
\label{sec:sc_dmd}

We distill a pre-trained teacher into a few-step student generator parameterized by a velocity field
$v_\theta(x_t,t,c)$ under the probability flow ODE view.
The student induces a one-step Euler transport operator
\begin{equation}
\Psi_{\theta}^{t\rightarrow s}(x)
~\triangleq~
x-(t-s)\,v_{\theta}(x,t,c),
\qquad 0\le s<t\le 1,
\label{eq:euler_map}
\end{equation}
and a $K$-step Euler sampler on a schedule $\mathcal{T}^{(K)}=\{t_0>\cdots>t_K\}$ iterates
$x_{t_{i+1}}=\Psi_{\theta}^{t_i\rightarrow t_{i+1}}(x_{t_i})$.
At inference, we consider both deterministic Euler integration and consistency-style sampling (CM solver)~\cite{song2023consistency}, which alternates denoising and re-noising and is widely used in modern autoregressive generation settings~\cite{yin2025slow,huang2025self}.
Since both solvers repeatedly apply the same student model, generation quality ultimately depends not only on the quality of each individual update, but also on how well these updates compose across timesteps.

\begin{figure}[t]
  \centering
  \includegraphics[width=0.9\columnwidth]{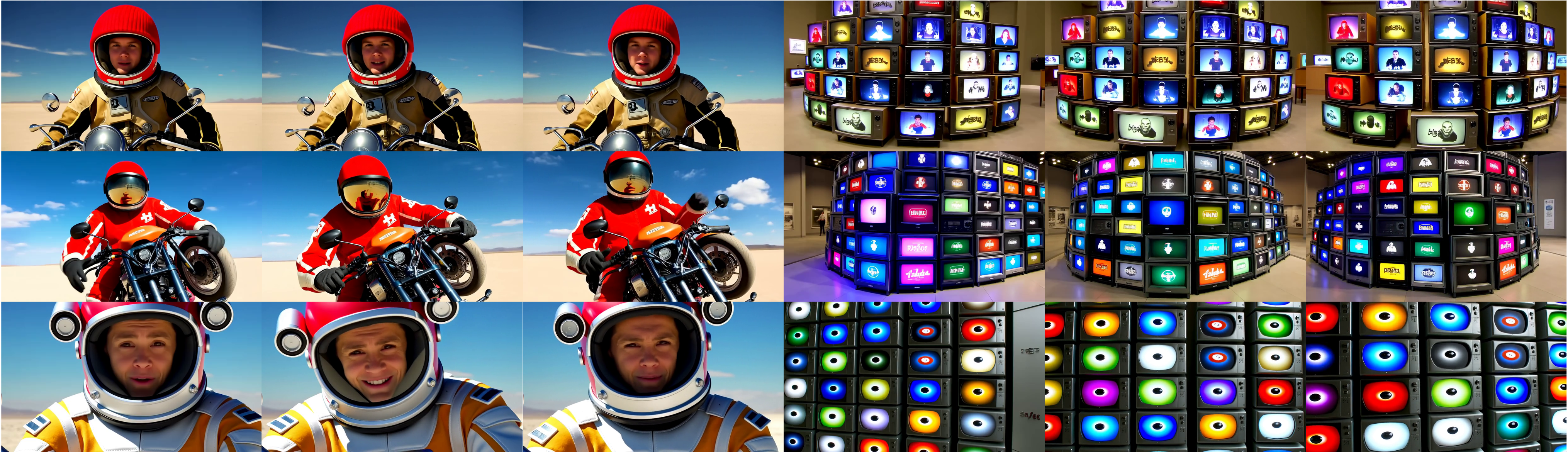}
  \caption{\textbf{Compositionality deficit of DMD.} First, middle, and last frames from 4-/8-/16-step DMD students (rows, top to bottom) on: (a) ``...a spaceman wearing a red wool knitted motorcycle helmet...'' and (b) ``...a large stack of vintage televisions all showing different programs...museum gallery.'' Increasing the number of denoising steps degrades rather than improves quality: the 16-step model loses the knitted helmet texture and corrupts the motorcycle structure, and produces incoherent television details.}
  \label{fig:dmd-error}
  \vspace{-0.5cm}
\end{figure}

\noindent\textbf{Compositionality deficit of DMD.}
DMD provides \emph{local} distribution-matching supervision at sampled noise levels (Eq.~\ref{eq:dmd_grad_prelim}), encouraging each trained timestep to act as a strong one-step generator.
However, few-step inference is inherently compositional: it relies on chaining the learned update operators across timesteps.
Even in recent DMD variants~\cite{huang2025self,yin2024improved} that use backward simulation to better approximate inference-time inputs, the training objective itself remains \emph{per-timestep and independent across noise levels}.
At timestep $t_i$, DMD only supervises the output distribution of the current step given input $x_{t_i}$; it does not explicitly penalize how neighboring updates interact under composition.
As a result, the learned operators can be individually well-calibrated yet collectively incoherent, leading to step-count-dependent behavior and unstable multi-step rollouts.

We empirically verify this failure mode in a clean, non-autoregressive setting.
Using the Wan~2.1 1.3B T2V backbone~\cite{wan2025wan}, we train three DMD students on 4-, 8-, and 16-timestep grids, respectively, each receiving distribution-matching supervision on its own set of noise levels using the standard backward-simulation strategy.
We then compare generation quality across these step counts using the same inputs~\cite{song2023consistency}.
As shown in Fig.~\ref{fig:dmd-error}, increasing the number of denoising steps does not improve generation quality; instead, longer rollouts exhibit artifacts such as over-exposure and semantic degradation.
This mirrors the degradation that AYF~\cite{sabouralign} formally proves can arise in consistency models under the same CM solver.
Our observation suggests that the issue is more fundamental: it stems from the absence of explicit compositional constraints in the learned denoising operators, rather than from a particular distillation objective.

\noindent\textbf{SC-DMD: shortcut self-consistency as semigroup-defect regularization.}
To address this issue, we augment DMD with a shortcut self-consistency regularizer.
For an exact flow map $\Phi^{t\rightarrow s}$, the semigroup law in Eq.~\ref{eq:semigroup} holds.
We therefore encourage the student-induced Euler operator $\Psi_\theta$ to approximately satisfy the same compositional property on the training grid.
Given a triple $(t_s,t_m,t_e)$ with $t_s>t_m>t_e$, we compute a \emph{direct (one-step) endpoint}
\begin{equation}
x^{(1)}_{t_e}
~=~
\Psi_{\theta}^{t_s\rightarrow t_e}(x_{t_s})
~=~
x_{t_s}-(t_s-t_e)\,v_{\theta}(x_{t_s},t_s,c),
\label{eq:one_step_end}
\end{equation}
and a \emph{composed (two-step) endpoint}
\begin{equation}
x^{(2)}_{t_e}
~=~
\Psi_{\theta}^{t_m\rightarrow t_e}\Big(\Psi_{\theta}^{t_s\rightarrow t_m}(x_{t_s})\Big).
\label{eq:two_step_end}
\end{equation}
The self-consistency (SC) loss then penalizes the discrepancy between these two endpoints:
\begin{equation}
\mathcal{L}_{\mathrm{SC}}
~=~
\mathbb{E}\Big[d\big(x^{(1)}_{t_e},x^{(2)}_{t_e}\big)\Big],
\qquad d(a,b)=\|a-b\|_2^2.
\label{eq:sc_loss}
\end{equation}
Intuitively, this regularizer enforces that a direct update from $t_s$ to $t_e$ agrees with the composed update $t_s\!\rightarrow\!t_m\!\rightarrow\!t_e$, thereby explicitly correcting the compositionality deficit identified above.

\noindent\emph{Grid and sampling design.}
Vanilla DMD typically trains on a coarse grid aligned with the $K$-step inference schedule, leaving little room to sample intermediate times $t_m$.
We instead train on a finer grid $\mathcal{T}_{\mathrm{train}}$ (e.g., $|\mathcal{T}_{\mathrm{train}}|{=}8$) to expose the model to richer intermediate-step structure.
For each $t_s\in\mathcal{T}_{\mathrm{train}}$, we sample the shortcut triple as
\begin{equation}
t_e \sim \mathcal{T}^{(K)} \cap \{t: t<t_s\}, \qquad
t_m \sim \mathcal{T}_{\mathrm{train}} \cap (t_e,t_s),
\label{eq:time_sampling}
\end{equation}
anchoring $t_e$ to the inference grid so that the shortcut constraint directly supervises the endpoints that matter at test time, while drawing $t_m$ from the finer training grid where dense DMD supervision is available.

Finally, we train the student with
\begin{equation}
\min_{\theta}\;\mathcal{L}_{\mathrm{DMD}}(\theta;\psi) + \lambda_{\mathrm{SC}}\mathcal{L}_{\mathrm{SC}}(\theta),
\label{eq:final_obj}
\end{equation}
where the critic parameters $\psi$ are updated by the standard DMD alternating procedure.
The SC term adds only lightweight overhead---one additional forward pass at $t_m$---yet explicitly couples the student's per-step operators into a more coherent multi-step trajectory.
As a result, it directly targets the failure mode in Fig.~\ref{fig:dmd-error}.
Ablations in Sec.~\ref{sec:ablation} confirm this effect: while vanilla DMD-8 suffers from quality degradation and dynamic collapse, SC-DMD achieves robust improvements on both metrics from 4- to 8-step inference.

\begin{figure}[t]
  \centering
  \includegraphics[width=0.9\columnwidth]{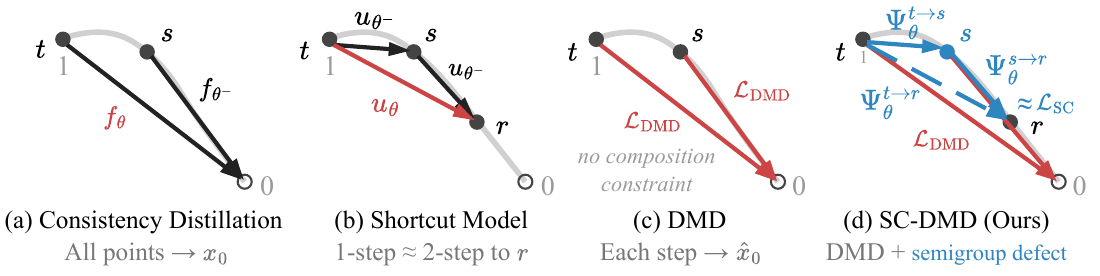}
  \caption{
  \textbf{Comparison of training trajectories for few-step distillation methods.}Consistency distillation, shortcut models, and flow-map distillation all impose composition-related constraints in different forms. SC-DMD addresses the compositionality deficit by introducing a semigroup-defect regularizer that aligns direct and composed updates while preserving DMD as the distribution matching objective.
  }
  \label{fig:scdmd}
  \vspace{-0.5cm}
\end{figure}

\begin{figure}[t]
  \centering
  \includegraphics[width=0.95\columnwidth]{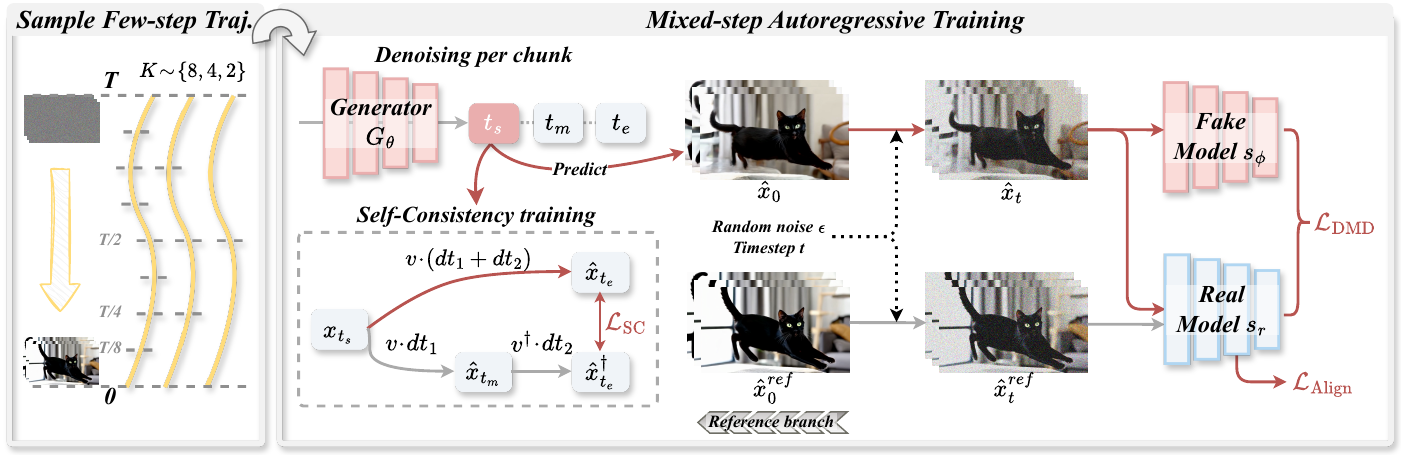}
  \caption{\textbf{Overview of \textit{Salt} for autoregressive video generation.}
  \emph{Left:} A step count $K\!\in\!\{8,4,2\}$ is sampled to define the few-step denoising trajectory.
  \emph{Middle:} Conditioned on the current KV cache, text, and noise, the student generator $G_\theta$ denoises the current chunk. A self-consistency (SC) loss $\mathcal{L}_{\mathrm{SC}}$ regularizes the endpoint discrepancy between a direct update and a composed two-step update.
  \emph{Right:} During mixed-step autoregressive training, the predicted clean sample is re-noised at a random timestep and optimized with the DMD loss $\mathcal{L}_{\mathrm{DMD}}$, while a higher-quality reference branch provides the cache-conditioned alignment loss $\mathcal{L}_{\mathrm{align}}$.}
  \label{fig:overview}
  \vspace{-0.5cm}
\end{figure}

\noindent\emph{Connection to shortcut models and flow map distillation.}
The self-consistency loss in Eq.~\eqref{eq:sc_loss} shares a common mathematical root with shortcut models~\cite{frans2024one} and continuous-time flow map distillation~\cite{sabouralign}: all of them encourage approximate semigroup composition of learned transport operators.
As illustrated in Fig.~\ref{fig:scdmd}, shortcut models parameterize the step size explicitly and use composition consistency as a self-distillation target, while AYF~\cite{sabouralign} enforces the semigroup law across continuous $(t,s)$ pairs as the primary distillation objective.
In contrast, SC-DMD uses the semigroup-defect penalty as a \emph{regularizer} inside a distribution-matching framework.
This distinction is important: when semigroup-style constraints are used as the sole training signal, they reduce to regression losses that can induce mode averaging under video dynamics.
By retaining DMD as the primary driver of distribution alignment, SC-DMD preserves mode-seeking fidelity while using compositional regularization to rectify the compositionality deficit.
We further analyze this design choice in Sec.~\ref{sec:supp_shortcut_comparison}, showing that the SC and DMD losses should act on an aligned training grid to avoid conflicting supervision.

\subsection{Cache-condition-aware Training for AR Distillation}
\label{sec:cache_aware}

In autoregressive video generation, each newly generated chunk is conditioned on the KV cache of previously generated chunks.
Unlike standard conditioning inputs (\eg text), the KV cache is itself model-generated, and its quality varies with the quality of prior rollouts.
In practice, higher-step rollouts tend to produce richer and more faithful caches, whereas lower-step rollouts produce noisier and more degraded ones.
As a result, autoregressive inference exposes the model to a range of cache qualities rather than a single conditioning regime.

This view reveals two practical challenges.
First, if training always uses a fixed step count $K$, the model only observes one cache-quality regime, leading to a train--inference mismatch when rollout quality changes.
Second, as autoregressive errors accumulate across chunks, the KV cache can progressively drift away from the conditioning states seen during training, causing semantic and quality degradation.

We address both challenges under a single principle:
\emph{systematically cover the range of cache qualities encountered during autoregressive training, while selectively imposing stronger regularization where long-horizon composition is most fragile.}
Concretely, we introduce two training mechanisms:
(i) mixed-step rollout, which exposes the model to diverse cache qualities and enables SC-DMD in the autoregressive setting, and
(ii) cache-conditioned reference alignment, which regularizes outputs across cache-quality levels.

\noindent\textbf{Mixed-step rollout: covering cache quality and enabling SC-DMD.}
To expose the generator to a broad range of cache qualities during autoregressive training, we adopt a mixed-step rollout strategy:
at each iteration, we sample $K \in \{2,4,8\}$ and run chunk-wise autoregressive rollout using the corresponding schedule $\mathcal{T}^{(K)}$.
This exposes both the generator and the critic to KV-cache patterns produced by different rollout fidelities, including the degraded caches that arise under low-$K$ settings.

Importantly, mixed-step rollout also provides the training scaffold for applying SC-DMD in the autoregressive setting.
We apply the standard DMD updates on generated samples for all sampled step counts, but activate the SC-DMD semigroup-defect regularizer only when $K{=}8$ is sampled.
This design targets the longest composition chain, where multi-step inconsistency is most pronounced and self-consistency regularization is most beneficial.
Thus, mixed-step rollout serves two roles simultaneously: it broadens the cache regimes seen during training, and it enables SC-DMD to regularize the high-step regime that matters most for stable autoregressive generation.

\noindent\textbf{Cache-conditioned reference alignment.}
To further improve robustness, we introduce a \emph{cache-conditioned reference alignment} loss that regularizes outputs under weaker cache conditioning toward those under a stronger reference cache.
Given the same input noise $\epsilon$, the generator's output quality is largely governed by the quality of the conditioning KV cache.
For rollouts sampled at $K\in\{2,4\}$, we construct a higher-quality reference using the next denser schedule, \ie $2\!\rightarrow\!4$ and $4\!\rightarrow\!8$, and align the resulting outputs in a TRD-style~\cite{zhang2026videorepa} relational feature space.
Formally, let
\[
z^{\mathrm{low}},\, z^{\mathrm{ref}} \in \mathbb{R}^{F\times S\times D}
\]
denote the intermediate features extracted from the lower-quality rollout and the higher-quality reference rollout, respectively, at a sampled noise level $t_f$, where $F$ is the number of frames, $S$ is the number of spatial tokens per frame, and $D$ is the channel dimension.
We first $\ell_2$-normalize the features along the channel dimension and compute, for each frame $f$, the spatial token relation matrices
\begin{equation}
R_f^{\mathrm{low}}=\bar z_f^{\mathrm{low}} \bigl(\bar z_f^{\mathrm{low}}\bigr)^\top,
\qquad
R_f^{\mathrm{ref}}=\bar z_f^{\mathrm{ref}} \bigl(\bar z_f^{\mathrm{ref}}\bigr)^\top,
\label{eq:relation_matrix}
\end{equation}
where $\bar z$ denotes the normalized feature and
$R_f^{(\cdot)}\in\mathbb{R}^{S\times S}$ encodes pairwise token similarities within frame $f$. We then penalize the discrepancy between the two relation matrices with a margin-relaxed objective:
\begin{equation}
\mathcal{L}_{\mathrm{align}}
=
\frac{1}{F}
\sum_{f=1}^{F}
\frac{1}{S^2}
\sum_{i=1}^{S}\sum_{j=1}^{S}
\Big[
\big|
R_f^{\mathrm{low}}(i,j)-R_f^{\mathrm{ref}}(i,j)
\big|
-\delta
\Big]_+,
\label{eq:align_loss}
\end{equation}
where $[\cdot]_+=\max(\cdot,0)$ and $\delta$ is a margin hyperparameter.
This loss encourages the lower-quality rollout to preserve the spatial relational structure of the stronger reference while ignoring small mismatches that are likely uninformative.

\noindent\textbf{Final autoregressive training objective.}
As shown in Fig.~\ref{fig:overview}, the complete training objective for autoregressive distillation is
\begin{equation}
\min_{\theta}\;
\mathcal{L}_{\mathrm{DMD}}(\theta;\psi)
+ \lambda_{\mathrm{SC}}\,\mathcal{L}_{\mathrm{SC}}(\theta)\,\mathbf{1}[K{=}8]
+ \lambda_{\mathrm{align}}\,\mathcal{L}_{\mathrm{align}}(\theta)\,\mathbf{1}[K{\in}\{2,4\}],
\label{eq:ar_obj}
\end{equation}
where $\mathcal{L}_{\mathrm{SC}}$ is applied at $K{=}8$ to regularize the longest composition chain, and $\mathcal{L}_{\mathrm{align}}$ is activated for low-step rollouts as a soft constraint toward stronger cache-conditioned references.
In practice, the reference alignment is applied stochastically for efficiency; detailed scheduling is deferred to the implementation details.

\section{Experimental Results}
\label{sec:experiments}

\subsection{Experimental Setup}
\label{sec:exp_setup}

We evaluate our method in both non-autoregressive and autoregressive settings.
For non-autoregressive distillation, we study two regimes: image-to-video (I2V) on Wan~2.1 I2V 14B~\cite{wan2025wan} and text-to-video (T2V) on Wan~2.1 T2V 1.3B, both under extremely low inference budgets.
For autoregressive real-time generation, we follow the training and evaluation protocols of recent real-time T2V paradigms and apply our method on top of representative backbones including Self Forcing~\cite{huang2025self}, LongLive~\cite{yang2025longlive}, and Causal Forcing~\cite{zhu2026causal}.
More implementation details are provided in the appendix.

\noindent\textbf{Baselines.}
In the non-autoregressive setting, we compare against representative few-step video distillation methods including PCM~\cite{wang2024phased}, DMD~\cite{yin2024one}, LightX2V~\cite{lightx2v}, and rCM~\cite{zheng2025large}.
Unless otherwise specified, methods are compared under the same inference budget.
For LightX2V, we follow the official evaluation protocol and report results from the released model.
In the autoregressive setting, we build upon strong real-time baselines including Self Forcing, LongLive, and Causal Forcing, and apply the full \textbf{Salt} framework (\textsc{SC-DMD} with cache-conditioned training) without modifying model architectures or inference pipelines.

\noindent\textbf{Benchmarks.}
For non-autoregressive I2V, we train PCM, DMD, and our method on the same proprietary internal I2V training set and evaluate on VBench-I2V from VBench++~\cite{huang2025vbench++}, reporting the overall I2V Score and Quality Score together with key dimensions such as Subject/Background Consistency, Motion Smoothness, Dynamic Degree, Imaging Quality, and Temporal Flickering.
For non-autoregressive and autoregressive T2V, we report VBench~\cite{huang2023vbench} scores on 5-second generation.
To assess long-horizon transfer, we additionally evaluate 30-second autoregressive rollouts on VBench-Long from VBench++~\cite{huang2025vbench++}.

\subsection{Non-autoregressive Few-Step Distillation}
\label{sec:exp_nonar}

\noindent\textbf{Image-to-Video on Wan~2.1 14B.}
Tab.~\ref{tab:vbench_i2v} reports VBench-I2V results under an extremely low inference budget of 4 NFEs.
Under the same training setting as DMD (8 H200 GPUs, 800 iterations), SC-DMD achieves the best \emph{I2V Score} ($93.90$) and \emph{Imaging Quality} ($72.16$), while substantially improving temporal coherence over DMD: \emph{Background Consistency} increases from $92.79$ to $95.97$, \emph{Motion Smoothness} from $97.99$ to $98.37$, and \emph{Temporal Flickering} from $95.21$ to $97.41$.
These gains indicate that semigroup-defect regularization improves multi-step compositional stability without sacrificing visual fidelity.
Compared with LightX2V, our method is also much more training-efficient, while remaining competitive or superior on most key metrics.
Although \emph{Dynamic Degree} decreases relative to DMD ($58.46 \rightarrow 52.85$), the simultaneous gains in smoothness, flicker suppression, and imaging quality suggest that SC-DMD favors more coherent motion rather than simply increasing optical-flow magnitude.

We also report an adversarial variant, Ours-$\alpha$, which adds an extra discriminator during training.
It further improves \emph{Quality Score} to $81.71$ and \emph{Dynamic Degree} to $68.13$, showing that SC-DMD is compatible with stronger perceptual objectives.
More implementation details are provided in the appendix.

\noindent\textbf{Text-to-Video on Wan~2.1 1.3B.}
We further evaluate SC-DMD in the non-autoregressive T2V setting (``Diffusion models'' group in Tab.~\ref{tab:quant_compare_open_models}).
Under the standard 4-NFE setting, SC-DMD improves over DMD on both \emph{Total Score} ($83.36$ vs.\ $82.78$) and \emph{Quality Score} ($84.76$ vs.\ $84.39$), while remaining competitive with rCM on \emph{Semantic Score}.
More importantly, the gain persists even at 2 NFEs: SC-DMD improves \emph{Total Score} from $82.41$ to $82.85$ and \emph{Quality Score} from $83.49$ to $84.06$ over DMD.
These results show that the benefit of compositional regularization remains effective even under more aggressive few-step compression.

\begin{table}[t]
    \centering
    \small
    \setlength{\tabcolsep}{4.2pt}
    \renewcommand{\arraystretch}{1.08}
    \caption{VBench-I2V results on Wan 2.1 14B under a 4-NFE inference budget. Higher is better ($\uparrow$).}
    \label{tab:vbench_i2v}
    \begin{adjustbox}{max width=0.98\linewidth}
    \begin{tabular}{l c c c c c c c c}
    \toprule
    \multirow{2}{*}{\textbf{Method}} & \multirow{2}{*}{\textbf{NFE}} 
    & \multicolumn{7}{c}{\textbf{VBench-I2V Evaluation Scores} $\uparrow$} \\
    \cmidrule(l){3-9}
    & 
    & \makecell{\textbf{I2V}\\\textbf{Score}}
    & \makecell{\textbf{Quality}\\\textbf{Score}}
    & \makecell{\textbf{Backg.}\\\textbf{Consist.}}
    & \makecell{\textbf{Motion}\\\textbf{Smooth.}}
    & \makecell{\textbf{Dynamic}\\\textbf{Degree}}
    & \makecell{\textbf{Imaging}\\\textbf{Quality}}
    & \makecell{\textbf{Temporal}\\\textbf{Flicker.}} \\
    \midrule

    \rowcolor{gray!12}
    \multicolumn{9}{l}{\textit{Consistency distillation}} \\
    PCM~\cite{wang2024phased} 
        & 8 & 93.63 & 78.52 & \textbf{97.34} & 98.24 & 30.98 & 70.42 & \textbf{97.67} \\
    \midrule

    \rowcolor{gray!12}
    \multicolumn{9}{l}{\textit{Distribution matching}} \\
    DMD~\cite{yin2024one} 
        & 4 & 93.09 & 78.89 & 92.79 & 97.99 & 58.46 & 70.35 & 95.21 \\
    LightX2V~\cite{lightx2v} 
        & 4 & 93.50 & \underline{80.92} & 95.87 & 97.89 & \underline{60.33} & 71.67 & 96.30 \\
    \midrule

    \textbf{Ours} 
        & 4 & \textbf{93.90} & 80.86 & \underline{95.97} & \textbf{98.37} & 52.85 & \textbf{72.16} & \underline{97.41} \\
    \textbf{Ours-$\alpha$} 
        & 4 & \underline{93.88} & \textbf{81.71} & 95.46 & \underline{98.30} & \textbf{68.13} & \underline{72.08} & 96.48 \\
    \bottomrule
    \end{tabular}
    \end{adjustbox}
    \vspace{-0.15cm}
\end{table}

\begin{table}[t]
\centering
\small
\setlength{\tabcolsep}{4.2pt}
\renewcommand{\arraystretch}{1.08}
\caption{VBench results for 5-second video generation with representative open-source models of similar parameter sizes and resolutions. Higher is better ($\uparrow$).}
\label{tab:quant_compare_open_models}
\begin{adjustbox}{max width=0.98\linewidth}
\begin{tabular}{l c c c c c c}
\toprule
\textbf{Model} & \textbf{\#Params} & \textbf{Resolution} & \textbf{NFE}
& \multicolumn{3}{c}{\textbf{VBench Scores} $\uparrow$} \\
\cmidrule(l){5-7}
& & & 
& \makecell[c]{\textbf{Total}}
& \makecell[c]{\textbf{Quality}}
& \makecell[c]{\textbf{Semantic}} \\
\midrule

\rowcolor{gray!12}
\multicolumn{7}{l}{\textit{Diffusion models}} \\
rCM~\cite{zheng2025large}
    & 1.3B & $832\times480$ & 4 & 82.73 & 83.65 & \textbf{79.04} \\
DMD~\cite{yin2024one}
    & 1.3B & $832\times480$ & 4 & 82.78 & 84.39 & 76.36 \\
    & 1.3B & $832\times480$ & 2 & 82.41 & 83.49 & 78.06 \\
\textbf{Ours (SC-DMD)}
    & 1.3B & $832\times480$ & 4 & \textbf{83.36} & \textbf{84.76} & 77.77 \\
    & 1.3B & $832\times480$ & 2 & \textbf{82.85} & \textbf{84.06} & 78.01 \\
\midrule

\rowcolor{gray!12}
\multicolumn{7}{l}{\textit{Autoregressive models}} \\
CausVid~\cite{yin2025slow}
    & 1.3B & $832\times480$ & 4 & 81.20 & 84.05 & 69.80 \\
Self Forcing~\cite{huang2025self}
    & 1.3B & $832\times480$ & 4 & 84.20 & 84.74 & \textbf{82.05} \\
\textbf{Ours - Self Forcing}
    & 1.3B & $832\times480$ & 4 & \textbf{84.47} & \textbf{85.27} & 81.28 \\
LongLive~\cite{yang2025longlive}
    & 1.3B & $832\times480$ & 4 & 84.40 & 85.12 & 81.53 \\
\textbf{Ours - LongLive}
    & 1.3B & $832\times480$ & 4 & \textbf{84.93} & \textbf{85.41} & \textbf{83.00} \\
Causal Forcing~\cite{zhu2026causal}
    & 1.3B & $832\times480$ & 4 & 84.62 & 85.41 & 81.47 \\
\textbf{Ours - Causal Forcing}
    & 1.3B & $832\times480$ & 4 & \textbf{85.08} & \textbf{85.96} & \textbf{81.59} \\
    & 1.3B & $832\times480$ & 2 & \underline{84.80} & \underline{85.63} & \underline{81.49} \\
\bottomrule
\end{tabular}
\end{adjustbox}
\vspace{-0.15cm}
\end{table}

\subsection{Autoregressive Real-Time Video Generation}
\label{sec:exp_ar_short}

\noindent\textbf{5-second short video generation.}
Tab.~\ref{tab:quant_compare_open_models}
reports VBench scores for single-prompt 5-second generation.
We apply our method to three representative autoregressive backbones: Self Forcing~\cite{huang2025self}, LongLive~\cite{yang2025longlive}, and Causal Forcing~\cite{zhu2026causal}, without changing their architectures or inference pipelines.
Under the standard 4-NFE setting, our method consistently improves \emph{Total Score} and \emph{Quality Score} across all three backbones.
On LongLive, \emph{Total Score} improves from $84.40$ to $84.93$ and \emph{Quality Score} from $85.12$ to $85.41$;
on Self Forcing, \emph{Total Score} improves from $84.20$ to $84.47$ and \emph{Quality Score} from $84.74$ to $85.27$;
on Causal Forcing, \emph{Total Score} improves from $84.62$ to $85.08$ and \emph{Quality Score} from $85.41$ to $85.96$.
The strongest semantic gain appears on LongLive ($81.53 \rightarrow 83.00$), while Causal Forcing also improves in \emph{Semantic Score} ($81.47 \rightarrow 81.59$).
As a complementary human-feedback-oriented evaluation, Salt also consistently improves VideoAlign~\cite{liu2026improving} and its motion-quality dimension across the three autoregressive backbones; detailed results are provided in Sec.~\ref{sec:supp_videoalign_motion}.

The benefit also extends to more aggressive inference budgets.
In the Causal Forcing family, our 2-step model reaches a \emph{Total Score} of $84.80$ and a \emph{Quality Score} of $85.63$, surpassing the original 4-step Causal Forcing baseline ($84.62$ Total, $85.41$ Quality).
This shows that our training objective remains effective even when the inference budget is further reduced.

\begin{figure}[t]
  \centering
  \includegraphics[width=1.0\columnwidth]{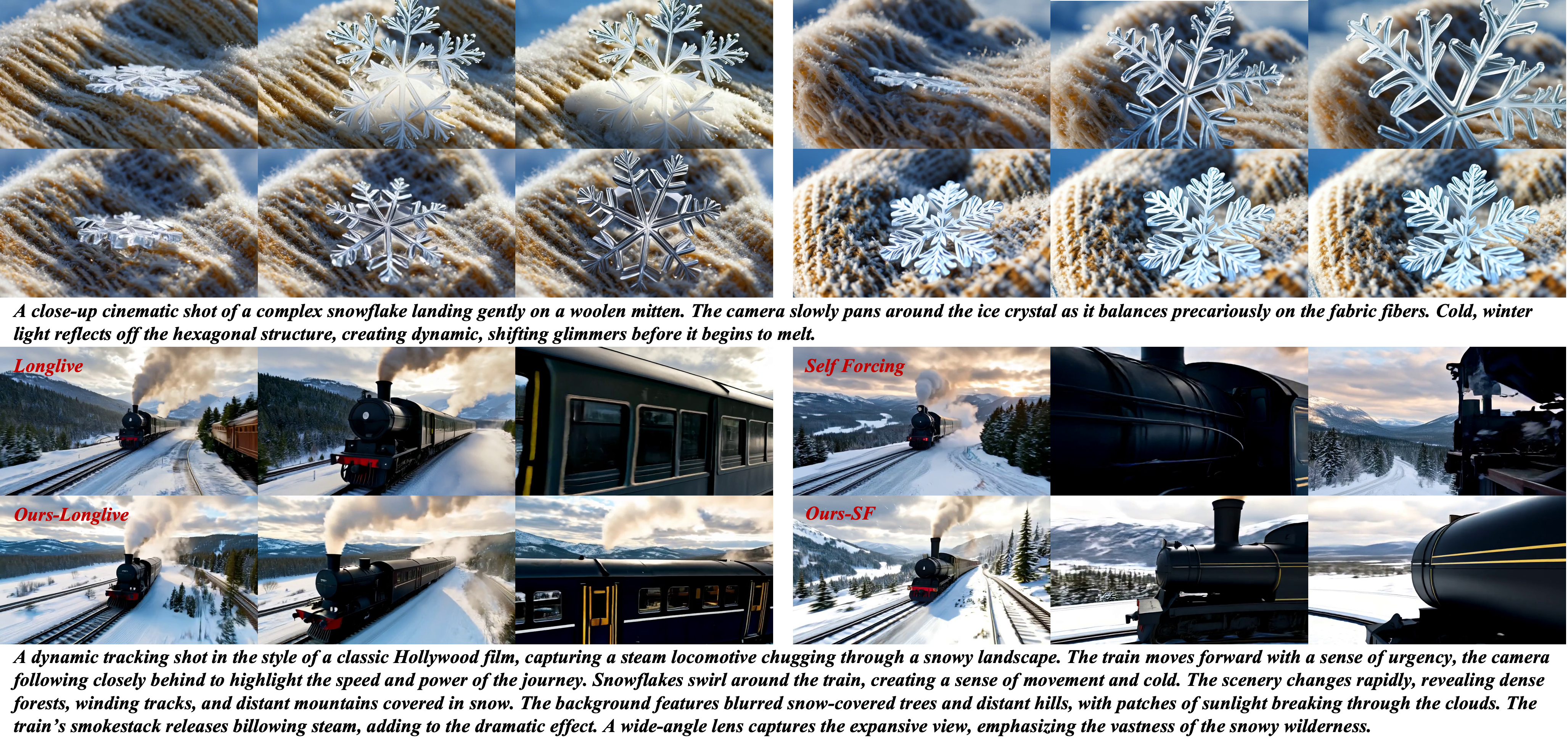}
  \caption{Qualitative comparison on texture-rich and high-dynamic scenes.}
  \label{fig:ar_vis}
  \vspace{-0.3cm}
\end{figure}

\noindent\textbf{Visualization results.}
Fig.~\ref{fig:ar_vis} shows two challenging cases: one dominated by fine-grained textures and the other by high-dynamic motion.
In the snowflake example, the LongLive baseline progressively loses the detailed wool texture and produces visible blurring, whereas Ours-LongLive preserves the fine-grained fabric structure.
In the train example, the Self Forcing baseline exhibits structural distortion under rapid motion, while Ours-SF maintains a coherent locomotive shape and more stable scene geometry.
These examples qualitatively support the gains in visual fidelity and temporal stability reported in Tab.~\ref{tab:quant_compare_open_models}.

\noindent\textbf{30-second long video generation.}
\label{sec:exp_ar_long}
Although our method is designed for real-time autoregressive generation rather than explicit long-video optimization, we evaluate whether its benefits transfer to longer horizons.
Following CausVid~\cite{yin2025slow}, we select 128 prompts from MovieGenBench and generate five 30-second videos per prompt, with Infinity-RoPE applied to both baselines and our method as a shared long-generation setup.
Results on VBench-Long are reported in Tab.~\ref{tab:vbench-long}.

The gains carry over to long-horizon generation.
On LongLive, our method improves \emph{Total Score} from $79.03$ to $79.27$, \emph{Quality Score} from $82.82$ to $82.90$, and \emph{Semantic Score} from $63.88$ to $64.74$.
On Causal Forcing, \emph{Total Score} improves from $78.11$ to $78.28$, while \emph{Semantic Score} increases substantially from $60.25$ to $62.77$.
These results suggest that the per-chunk compositional gains from SC-DMD and the broader cache exposure from mixed-step training transfer beyond short-horizon rollout, especially in overall quality and semantic consistency.

\begin{table*}[t]
\centering
\small
\setlength{\tabcolsep}{4.2pt}
\renewcommand{\arraystretch}{1.08}
\caption{VBench-Long results for 30-second autoregressive video generation. Higher is better ($\uparrow$). \textbf{Bold} indicates the best within each family.}
\label{tab:vbench-long}
\begin{adjustbox}{max width=\textwidth}
\begin{tabular}{l c c c c c c c c c}
\toprule
\textbf{Model}
& \multicolumn{9}{c}{\textbf{VBench-Long Evaluation Scores} $\uparrow$} \\
\cmidrule(l){2-10}
&
\shortstack[c]{\textbf{Total}\\\textbf{Score}}
&
\shortstack[c]{\textbf{Quality}\\\textbf{Score}}
&
\shortstack[c]{\textbf{Semantic}\\\textbf{Score}}
&
\shortstack[c]{\textbf{Subject}\\\textbf{Consist.}}
&
\shortstack[c]{\textbf{Temp.}\\\textbf{Flicker.}}
&
\shortstack[c]{\textbf{Motion}\\\textbf{Smooth.}}
&
\shortstack[c]{\textbf{Dynamic}\\\textbf{Degree}}
&
\shortstack[c]{\textbf{Aesthetic}\\\textbf{Quality}}
&
\shortstack[c]{\textbf{Imaging}\\\textbf{Quality}} \\
\midrule

\rowcolor{gray!12}
\multicolumn{10}{l}{\textit{Causal Forcing family}} \\
Causal Forcing~\cite{zhu2026causal}
& 78.11 & \textbf{82.57} & 60.25 & 96.29 & 95.47 & 97.67 & \textbf{76.95} & 56.50 & 71.12 \\
\textbf{Ours - Causal Forcing}
& \textbf{78.28} & 82.15 & \textbf{62.77} & \textbf{97.16} & \textbf{96.95} & \textbf{98.32} & 54.02 & \textbf{57.42} & 70.99 \\
\midrule

\rowcolor{gray!12}
\multicolumn{10}{l}{\textit{LongLive family}} \\
LongLive~\cite{yang2025longlive}
& 79.03 & 82.82 & 63.88 & \textbf{97.36} & 96.62 & 98.16 & 63.45 & 58.26 & \textbf{70.90} \\
\textbf{Ours - LongLive}
& \textbf{79.27} & \textbf{82.90} & \textbf{64.74} & 97.28 & \textbf{96.92} & \textbf{98.29} & \textbf{63.68} & \textbf{58.35} & 70.07 \\
\bottomrule
\end{tabular}
\end{adjustbox}
\end{table*}

\subsection{Ablation Study}
\label{sec:ablation}

\noindent\textbf{Effect of $\mathcal{L}_{\mathrm{SC}}$: fixing the compositionality deficit.}
We first ablate SC-DMD in a clean non-autoregressive setting on Wan~2.1 1.3B T2V, where all variants use the same 4-step inference budget.
Tab.~\ref{tab:ablation_nonar} compares three students: DMD-4, trained on a 4-point grid; DMD-8, trained on a denser 8-point grid; and SC-DMD, which uses the same 8-point grid as DMD-8 but adds the shortcut self-consistency loss.
Simply increasing the training grid density does not solve the problem: compared with DMD-4, DMD-8 drops from $84.39$ to $84.05$ in \emph{Quality} and from $82.78$ to $82.54$ in \emph{Total}, while also degrading several structure-sensitive dimensions such as \emph{Spatial Relation}, \emph{Multi-Objects}, \emph{Object Class}, and \emph{Imaging Quality}.
In contrast, SC-DMD consistently improves over both DMD baselines, achieving the best \emph{Total} ($83.36$), \emph{Quality} ($84.76$), and \emph{Semantic} ($77.77$), together with clear gains on the fine-grained structural dimensions.
This confirms that the gain comes from explicit compositional regularization rather than from training on a denser timestep grid.

\begin{table*}[t]
\centering
\small
\setlength{\tabcolsep}{4.2pt}
\renewcommand{\arraystretch}{1.08}
\caption{Ablation of SC-DMD on Wan~2.1 1.3B T2V under 4-step inference. Increasing the DMD training grid from 4 to 8 timesteps does not improve few-step generation, whereas SC-DMD consistently improves both overall and structure-sensitive metrics. Higher is better ($\uparrow$).}
\label{tab:ablation_nonar}
\begin{adjustbox}{max width=\textwidth}
\begin{tabular}{l c c c c c c c}
\toprule
\textbf{Method}
& \makecell[c]{\textbf{Quality}}
& \makecell[c]{\textbf{Semantic}}
& \makecell[c]{\textbf{Total}}
& \makecell[c]{\textbf{Spatial}\\\textbf{Rel.}}
& \makecell[c]{\textbf{Multi-}\\\textbf{Objects}}
& \makecell[c]{\textbf{Object}\\\textbf{Class}}
& \makecell[c]{\textbf{Imaging}\\\textbf{Q.}} \\
\midrule
DMD-8        & 84.05 & 76.50 & 82.54 & 67.13 & 78.98 & 93.47 & 65.46 \\
DMD-4        & 84.39 & 76.36 & 82.78 & 69.49 & 79.37 & 93.23 & 64.73 \\
\textbf{SC-DMD (Ours)} & \textbf{84.76} & \textbf{77.77} & \textbf{83.36} & \textbf{71.91} & \textbf{83.89} & \textbf{93.91} & \textbf{67.32} \\
\bottomrule
\end{tabular}
\end{adjustbox}
\vspace{-0.2cm}
\end{table*}

\begin{figure}[t]
  \centering
  \includegraphics[width=0.95\columnwidth]{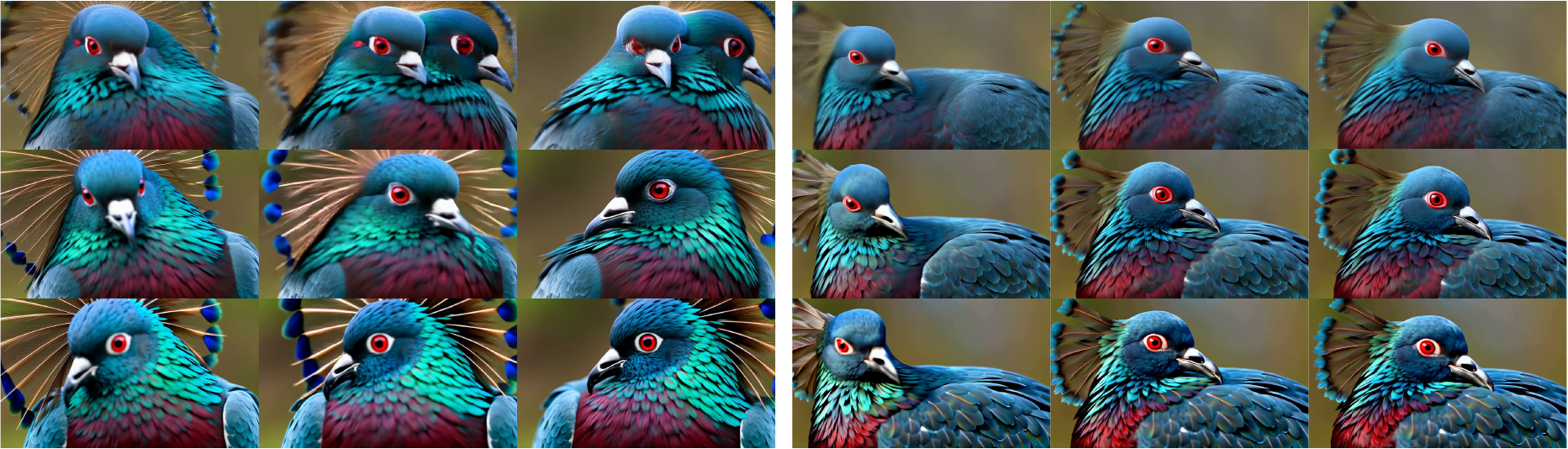}
  \caption{\textbf{Cross-step consistency under the same seed and prompt.} Left: DMD. Right: SC-DMD. Rows correspond to 2-, 4-, and 8-step sampling, respectively. Vanilla DMD exhibits strong step-dependent variation and structural corruption as the number of steps changes, whereas SC-DMD yields much more consistent outputs across step counts and better preserves object structure.}
  \label{fig:ablation_consistency}
  \vspace{-0.2cm}
\end{figure}

\noindent\textbf{Cross-step compositional consistency.}
Fig.~\ref{fig:ablation_consistency} compares DMD and SC-DMD under the same prompt and random seed across 2-, 4-, and 8-step sampling.
Vanilla DMD exhibits clear step-dependent inconsistency: the generated content changes noticeably as the step count varies, and higher-step sampling can introduce structural artifacts such as duplicated heads and unstable object identity.
In contrast, SC-DMD produces much more consistent outputs across 2/4/8 steps and is markedly less prone to structural corruption.
This qualitative evidence directly supports our central claim: SC-DMD improves the \emph{composition} of learned updates, rather than merely optimizing one particular operating point.

\begin{table}[t]
\centering
\small
\setlength{\tabcolsep}{4.2pt}
\renewcommand{\arraystretch}{1.08}
\caption{Ablation of autoregressive training components on Causal Forcing under the VBench-extend prompt setting. The 4-NFE block shows the main component ablation; the 2-NFE block shows low-budget transfer, where reference alignment becomes more helpful. Higher is better ($\uparrow$).}
\label{tab:ablation_ar}
\begin{adjustbox}{max width=0.98\linewidth}
\begin{tabular}{l c c c c}
\toprule
\textbf{Method} & \textbf{NFE} & \textbf{Quality} & \textbf{Semantic} & \textbf{Total} \\
\midrule
\rowcolor{gray!12}
\multicolumn{5}{l}{\textit{4-NFE main ablation}} \\
Causal Forcing (official baseline)               & 4 & 85.41 & 81.47 & 84.62 \\
Causal Forcing + naive $\mathcal{L}_{\mathrm{SC}}$       & 4 & 84.35 & 81.77 & 83.83 \\
Causal Forcing + mixed-step + $\mathcal{L}_{\mathrm{SC}}$ & 4 & 85.91 & 81.48 & 85.02 \\
\textbf{Ours (full)}                             & 4 & \textbf{85.96} & \textbf{81.59} & \textbf{85.08} \\
\midrule
\rowcolor{gray!12}
\multicolumn{5}{l}{\textit{2-NFE low-budget transfer}} \\
Causal Forcing + mixed-step + $\mathcal{L}_{\mathrm{SC}}$ & 2 & 85.62 & 80.65 & 84.63 \\
\textbf{Ours (full)}                             & 2 & \textbf{85.63} & \textbf{81.49} & \textbf{84.80} \\
\bottomrule
\end{tabular}
\end{adjustbox}
\vspace{-0.1cm}
\end{table}

\noindent\textbf{Ablation of autoregressive training components.}
We next ablate the autoregressive training recipe on Causal Forcing under the same VBench-extend prompt setting as the main AR results.
Tab.~\ref{tab:ablation_ar} shows three key observations.
First, naively adding $\mathcal{L}_{\mathrm{SC}}$ to the official baseline hurts performance (\emph{Total}: $84.62 \rightarrow 83.83$), showing that self-consistency alone is not sufficient in the autoregressive setting.
Second, once paired with mixed-step rollout, SC-DMD becomes effective and improves the 4-NFE baseline to $85.02$ in \emph{Total}, confirming that mixed-step rollout is the training scaffold that makes SC regularization work in AR generation.
Third, the full model achieves the best 4-NFE result ($85.08$ \emph{Total}, $81.59$ \emph{Semantic}) after adding reference alignment.
Notably, the gain from reference alignment becomes more pronounced at 2 NFEs: compared with the mixed-step+$\mathcal{L}_{\mathrm{SC}}$ variant, the full model improves \emph{Semantic} from $80.65$ to $81.49$ and \emph{Total} from $84.63$ to $84.80$.
This is consistent with the intended role of the alignment term, namely guiding weaker low-step rollouts toward stronger high-step references.

\section{Conclusion}
We presented \textbf{Salt}, a training framework for few-step video diffusion distillation that addresses two structural bottlenecks.
First, we identified the \emph{compositionality deficit} of DMD: its purely local supervision does not enforce coherent multi-step generation.
To remedy this, we proposed \textbf{SC-DMD}, which builds on distribution matching with a lightweight semigroup-defect regularizer that encourages endpoint-consistent composition of denoising updates.
Second, for autoregressive generation, we introduced \emph{cache-conditioned training} to handle the quality variability of model-generated KV caches: mixed-step rollout exposes the model to diverse cache regimes during training, while reference alignment regularizes weak-cache outputs toward stronger references.
Across both non-autoregressive (Wan~2.1) and autoregressive (Self Forcing, Causal Forcing, and LongLive) settings, \textbf{Salt} delivers consistent gains in few-step generation quality, cross-step compositional consistency, and long-horizon semantic stability.

\section*{Acknowledgement}
This work was supported by the Hong Kong Research Grants Council under the Areas of Excellence scheme grant AoE/E-601/22-R and NSFC/RGC Collaborative Research Scheme grant CRS\_HKUST603/22.


\bibliographystyle{splncs04}
\bibliography{main}

\newpage
\appendix 
\setcounter{table}{0}
\setcounter{figure}{0}
\renewcommand{\theHsection}{appendix.\Alph{section}}
\renewcommand{\theHtable}{appendix.\arabic{table}}
\renewcommand{\theHfigure}{appendix.\arabic{figure}}

\section{More Results and Analysis}

\subsection{Measuring Semigroup Defect on the Test-Time Inference Path}
\label{sec:supp_local_semigroup_defect}

A key motivation of SC-DMD is that vanilla DMD provides supervision at individual noise levels, but does not explicitly constrain whether the learned timestep-to-timestep operators compose coherently across a multi-step inference path. Since our method is designed to reduce this \emph{compositionality deficit}, we provide a direct diagnostic by measuring the \emph{local semigroup defect} of the learned student operator on the test-time 4-step inference schedule.

\paragraph{Definition.}
Using the notation in Sec.~4.1, for a shortcut triple $(t_s, t_m, t_e)$ with $t_s > t_m > t_e$, we define the direct endpoint
\begin{equation}
x_{t_e}^{(1)} = \Psi_{\theta}^{t_s \rightarrow t_e}(x_{t_s}),
\end{equation}
and the composed endpoint
\begin{equation}
x_{t_e}^{(2)} = \Psi_{\theta}^{t_m \rightarrow t_e}\!\left(\Psi_{\theta}^{t_s \rightarrow t_m}(x_{t_s})\right).
\end{equation}
Their discrepancy is exactly the quantity regularized by the SC loss in Eq.~(8). To analyze this discrepancy at test time, we define the following displacement-normalized local semigroup defect:
\begin{equation}
\Delta_{\mathrm{sg}}(t_s,t_m,t_e)
=
\mathbb{E}_{x_{t_s}}
\left[
\frac{
\left\|
x_{t_e}^{(1)} - x_{t_e}^{(2)}
\right\|_2^2
}{
\left\|
x_{t_e}^{(1)} - x_{t_s}
\right\|_2^2 + \epsilon
}
\right].
\label{eq:supp_local_semigroup_defect}
\end{equation}

\paragraph{Why displacement normalization.}
A naive normalization by the endpoint latent norm can be misleading for low-displacement or partially degenerate operators: if both the direct and composed trajectories move very little, their discrepancy may be small even when the operator is not meaningfully compositional. In contrast, Eq.~\eqref{eq:supp_local_semigroup_defect} normalizes the discrepancy by the magnitude of the direct denoising transport $\|x_{t_e}^{(1)} - x_{t_s}\|_2^2$, and therefore measures composition error \emph{relative to the amount of transport actually performed}.

\begin{figure}[t]
    \centering
    \includegraphics[width=0.6\linewidth]{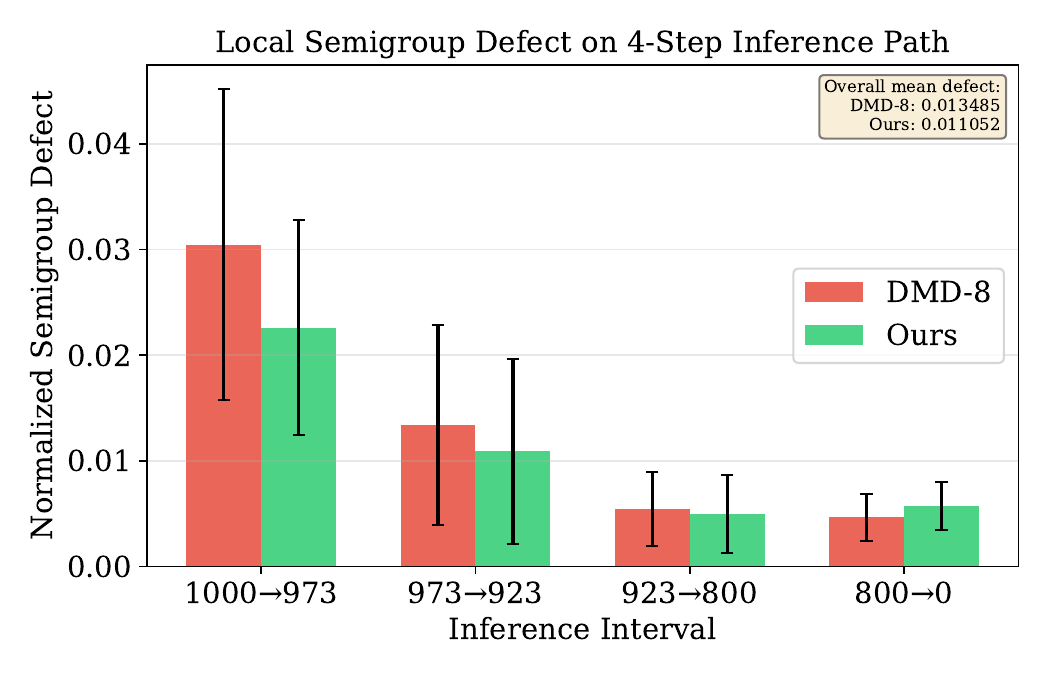}
    \caption{
    Displacement-normalized local semigroup defect on the test-time 4-step inference path.
    For each adjacent inference interval $(t_s,t_e)$, we compare the direct endpoint $x_{t_e}^{(1)}=\Psi_{\theta}^{t_s \rightarrow t_e}(x_{t_s})$ against the composed endpoint $x_{t_e}^{(2)}=\Psi_{\theta}^{t_m \rightarrow t_e}(\Psi_{\theta}^{t_s \rightarrow t_m}(x_{t_s}))$, where $t_m$ is the corresponding intermediate timestep from the finer training grid.
    Lower is better.
    SC-DMD achieves a lower overall local semigroup defect than the DMD baseline, supporting the claim that self-consistency regularization improves the compositional behavior of the learned denoising operator.
    }
    \label{fig:supp_semigroup_defect}
\end{figure}

\paragraph{Evaluation protocol.}
We evaluate Eq.~\eqref{eq:supp_local_semigroup_defect} along the 4-step inference schedule used at test time. In our implementation, this inference schedule is generated using a timestep shift of 12, which gives rise to the four adjacent inference intervals shown in Fig.~\ref{fig:supp_semigroup_defect}. For each adjacent inference interval $(t_s,t_e)$ on $\mathcal{T}^{(K)}$, we instantiate the shortcut triple following the same design as Eq.~(9): $t_e$ is anchored to the inference grid, and $t_m$ is taken from the finer training grid between $t_s$ and $t_e$. We then compute the defect for each interval and average over evaluation samples.

\paragraph{Results.}
Fig.~\ref{fig:supp_semigroup_defect} shows the resulting local semigroup defect on the test-time inference path. Overall, SC-DMD achieves a lower displacement-normalized local semigroup defect than the DMD baseline (0.0111 vs.\ 0.0135). The improvement is most evident on the early and middle inference intervals, while the final low-noise interval exhibits a smaller gap. This trend provides direct mechanistic support for our central claim: beyond improving downstream video quality, the proposed self-consistency regularization also makes the learned student operator more coherent under local composition on the test-time denoising path.

\paragraph{Interpretation.}
We emphasize that this metric is a \emph{local} diagnostic of compositional behavior rather than a replacement for rollout-level generation benchmarks. Nonetheless, its trend is consistent with the main-paper observation that SC-DMD improves few-step generation quality while mitigating the degradation associated with multi-step composition.

\subsection{Comparison with Shortcut-Style Consistency}
\label{sec:supp_shortcut_comparison}

We further compare SC-DMD with a shortcut-style flow-matching objective to clarify the role of self-consistency.
Tab.~\ref{tab:supp_shortcut_comparison} shows that SCFM~\cite{cai2025shortcutting} substantially underperforms DMD-4 on Wan~2.1 T2V under the same 4-NFE evaluation setting, and a naive DMD+SCFM hybrid further degrades performance.
In contrast, SC-DMD obtains the best overall result by keeping DMD as the primary distribution-matching objective and using shortcut self-consistency only as a semigroup-defect regularizer.
The degradation of the naive hybrid is not merely due to adding an extra regularizer; it also stems from a mismatch between the shortcut training grid and the DMD distillation grid.
In the naive DMD+SCFM setting, shortcut supervision is applied on a dense shortcut grid, whereas DMD supervises only a coarse set of distillation timesteps.
This creates unevenly optimized operators: some timesteps receive both distribution matching and shortcut consistency, while others receive only shortcut-style trajectory supervision.
Consequently, the composed branch may pass through under-optimized intermediate operators, making the two-step endpoint an unreliable target for the direct update.
This grid mismatch induces an objective conflict between distribution matching and shortcut consistency, and can amplify trajectory non-smoothness when the unreliable composed endpoint is recursively reused as a consistency target.
SC-DMD avoids this issue by applying the SC regularizer on the same DMD-supervised training grid, so the consistency constraint regularizes already distribution-matched operators instead of propagating shortcut-only errors.
The sensitivity study shows that larger $\lambda_{\mathrm{SC}}$ values remain competitive but shift the trade-off toward semantic score at the cost of quality; our default $\lambda_{\mathrm{SC}}{=}0.2$ gives the best total score.

\begin{table}[H]
\centering
\small
\setlength{\tabcolsep}{4.5pt}
\renewcommand{\arraystretch}{1.05}
\caption{Additional analysis of shortcut-style consistency on Wan~2.1 T2V under 4-NFE inference. SC-DMD keeps DMD as the main distribution-matching objective and uses self-consistency as a regularizer. Higher is better ($\uparrow$).}
\label{tab:supp_shortcut_comparison}
\begin{adjustbox}{max width=0.98\linewidth}
\begin{tabular}{l c c c}
\toprule
\textbf{Method} & \textbf{Quality} & \textbf{Semantic} & \textbf{Total} \\
\midrule
\rowcolor{gray!12}
\multicolumn{4}{l}{\textit{SCFM vs. SC-DMD}} \\
SCFM~\cite{cai2025shortcutting} & 79.47 & 70.63 & 77.70 \\
Naive DMD+SCFM & 77.17 & 66.64 & 75.06 \\
DMD-4 & 84.39 & 76.36 & 82.78 \\
\textbf{SC-DMD} ($\lambda_{\mathrm{SC}}{=}0.2$) & \textbf{84.76} & 77.77 & \textbf{83.36} \\
\midrule
\rowcolor{gray!12}
\multicolumn{4}{l}{\textit{Sensitivity to $\lambda_{\mathrm{SC}}$}} \\
SC-DMD ($\lambda_{\mathrm{SC}}{=}0.4$) & 84.40 & 78.11 & 83.27 \\
SC-DMD ($\lambda_{\mathrm{SC}}{=}1.0$) & 84.21 & \textbf{78.29} & 83.02 \\
\bottomrule
\end{tabular}
\end{adjustbox}
\vspace{-0.2cm}
\end{table}

\subsection{Additional Results on Standard VBench Prompts}
\label{sec:supp_t2v_results}

In the main paper, our autoregressive results are reported on the prompt-extended evaluation setting, following prior real-time video generation works such as Self Forcing, Causal Forcing, and LongLive for fair comparison.
Since the original VBench prompts are substantially shorter, we additionally evaluate our method on the \emph{standard} VBench prompt set to examine whether the gains transfer beyond the prompt-expansion protocol used in the main paper.

The results are shown in Tab.~\ref{tab:t2v_supp}.
For non-autoregressive text-to-video generation, we compare SC-DMD against rCM~\cite{zheng2025large} on Wan~2.1 T2V 1.3B~\cite{wan2025wan}, using the same 1.3B backbone, the same output resolution ($832 \times 480$), and the same CM solver.
SC-DMD consistently outperforms rCM on all three VBench metrics, improving \emph{Total} from $78.48$ to $80.72$, \emph{Quality} from $82.99$ to $84.41$, and \emph{Semantic} from $60.42$ to $65.97$.
This shows that the benefit of semigroup-based compositional regularization is not tied to the prompt-expanded setting and remains effective on the original short-prompt benchmark.

We also evaluate the autoregressive models on the same standard VBench prompts.
Although the absolute scores are lower than those under prompt extension, our method still yields consistent gains across all three backbones.
In particular, Ours-Self Forcing improves \emph{Total} from $82.16$ to $82.43$ and \emph{Quality} from $84.89$ to $85.29$;
Ours-Causal Forcing improves \emph{Total} from $82.28$ to $82.52$ and \emph{Semantic} from $70.35$ to $70.89$;
and Ours-LongLive improves \emph{Total} from $82.31$ to $82.60$ and \emph{Semantic} from $71.16$ to $71.30$.
These results suggest that the gains of \textbf{Salt} are robust to prompt style and generalize from the prompt-extended evaluation used in the main paper to the original standard VBench prompts.




\begin{table}[t]
\centering
\caption{Results on the standard VBench prompt set. The main paper reports prompt-extended evaluation for fair comparison with prior real-time video generation works; here we additionally report performance on the original short-prompt benchmark. Higher is better ($\uparrow$).}
\label{tab:t2v_supp}
\resizebox{\linewidth}{!}{
\begin{tabular}{
    >{\raggedright\arraybackslash}m{3.6cm}
    >{\centering\arraybackslash}m{1.6cm}
    >{\centering\arraybackslash}m{1.8cm}
    >{\centering\arraybackslash}m{1.2cm}
    *{3}{>{\centering\arraybackslash}m{1.4cm}}
}
\toprule
\multirow[c]{2}{*}{\textbf{Model}} &
\multirow[c]{2}{*}{\textbf{\#Params}} &
\multirow[c]{2}{*}{\textbf{Resolution}} &
\multirow[c]{2}{*}{\textbf{NFE}} &
\multicolumn{3}{c}{\textbf{VBench scores} $\uparrow$} \\
\cmidrule(l){5-7}
& & & & \textbf{Total} & \textbf{Quality} & \textbf{Semantic} \\
\midrule

\rowcolor{gray!15}
\multicolumn{7}{l}{\textit{Diffusion models}} \\
rCM~\cite{zheng2025large}          & 1.3B & $832\times480$ & 4 & 78.48 & 82.99 & 60.42 \\
Ours (SC-DMD)                      & 1.3B & $832\times480$ & 4 & \textbf{80.72} & \textbf{84.41} & \textbf{65.97} \\


\rowcolor{gray!15}
\multicolumn{7}{l}{\textit{Autoregressive models}} \\
CausVid~\cite{yin2025slow}        & 1.3B & $832\times480$ & 4 & 77.89 & 80.89 & 65.85 \\
Self Forcing~\cite{huang2025self} & 1.3B & $832\times480$ & 4 & 82.16 & 84.89 & 71.26 \\ 
\textbf{Ours - Self Forcing}   & 1.3B & $832\times480$ & 4 & \textbf{82.43} & \textbf{85.29} & \textbf{71.03} \\ 
Causal Forcing~\cite{zhu2026causal}                 & 1.3B & $832\times480$ & 4 & 82.28 & 85.27 & 70.35 \\
\textbf{Ours - Causal Forcing} & 1.3B & $832\times480$ & 4 & \textbf{82.52} & \textbf{85.43} & \textbf{70.89} \\
LongLive~\cite{yang2025longlive}      & 1.3B & $832\times480$ & 4 & 82.31 & 85.10 & 71.16 \\
\textbf{Ours - LongLive} & 1.3B & $832\times480$ & 4 & \textbf{82.60} & \textbf{85.42} & \textbf{71.30}  \\

\bottomrule
\end{tabular}
}
\end{table}

\subsection{Additional VideoAlign Evaluation and Motion Analysis}
\label{sec:supp_videoalign_motion}

We additionally evaluate the autoregressive models with VideoAlign~\cite{liu2026improving}, a human-feedback-oriented video reward model.
Tab.~\ref{tab:supp_videoalign} reports both the overall VideoAlign score and its motion-quality (MQ) dimension on the same prompt-extended 5-second setting used in the main autoregressive evaluation.
The improvements are consistent across all three backbones: VideoAlign increases by $+0.73$, $+0.54$, and $+1.07$ on Self Forcing, LongLive, and Causal Forcing, respectively; MQ also improves by $+0.41$, $+0.24$, and $+0.77$.
These results provide an independent check that Salt improves human-aligned video quality and motion plausibility beyond the VBench aggregate scores.

\begin{table}[t]
\centering
\small
\setlength{\tabcolsep}{3.8pt}
\renewcommand{\arraystretch}{1.05}
\caption{Additional VBench-extend and VideoAlign evaluation for 5-second autoregressive generation. MQ denotes the motion-quality dimension from VideoAlign. Higher is better ($\uparrow$).}
\label{tab:supp_videoalign}
\begin{adjustbox}{max width=0.98\linewidth}
\begin{tabular}{l c c c c c}
\toprule
\textbf{Model} & \textbf{Total} & \textbf{Quality} & \textbf{Semantic} & \textbf{MQ} & \textbf{VideoAlign} \\
\midrule
Self Forcing~\cite{huang2025self} & 84.20 & 84.74 & \textbf{82.05} & 1.69 & 7.75 \\
\textbf{Ours - Self Forcing} & \textbf{84.47} & \textbf{85.27} & 81.28 & \textbf{2.10} & \textbf{8.48} \\
\midrule
LongLive~\cite{yang2025longlive} & 84.40 & 85.12 & 81.53 & 1.76 & 8.04 \\
\textbf{Ours - LongLive} & \textbf{84.93} & \textbf{85.41} & \textbf{83.00} & \textbf{2.00} & \textbf{8.58} \\
\midrule
Causal Forcing~\cite{zhu2026causal} & 84.62 & 85.41 & 81.47 & 1.43 & 7.77 \\
\textbf{Ours - Causal Forcing} & \textbf{85.08} & \textbf{85.96} & \textbf{81.59} & \textbf{2.20} & \textbf{8.84} \\
\textbf{Ours - Causal Forcing} (2 NFE) & \underline{84.80} & \underline{85.63} & \underline{81.49} & -- & -- \\
\bottomrule
\end{tabular}
\end{adjustbox}
\vspace{-0.2cm}
\end{table}

This also helps interpret the Dynamic Degree trade-off in VBench-Long.
Dynamic Degree is based on optical-flow magnitude, so a larger value does not always imply better motion: flicker, abrupt scene changes, or noisy transitions can increase flow even when perceptual motion quality degrades.
For example, in the Causal Forcing family of the VBench-Long results, our Dynamic Degree decreases from $76.95$ to $54.02$, while \emph{Semantic Score}, \emph{Subject Consistency}, \emph{Temporal Flickering}, and \emph{Motion Smoothness} all improve.
The independent VideoAlign MQ gains in Tab.~\ref{tab:supp_videoalign} further support this interpretation: Salt improves motion quality on all three autoregressive backbones, indicating that lower optical-flow magnitude should not be read as worse motion by itself.
A concrete example is shown in the third group of Fig.~\ref{fig:supp_more_vis}.
In this Causal Forcing case, the baseline has a larger mean optical-flow magnitude than ours ($18.66$ vs.\ $11.50$), with a peak outlier of $209$ caused by an abrupt scene/identity switch: the main human subject suddenly changes across frames.
Such discontinuity can inflate the optical-flow-based Dynamic Degree even though the perceived motion is implausible.
By contrast, our result preserves the subject and scene continuity without the sudden switch, providing a qualitative counterpart to the higher VideoAlign MQ scores.

\section{Implementation Details}

\noindent\textbf{Non-autoregressive distillation.}
We evaluate SC-DMD in two non-AR settings: \emph{text-to-video} using Wan~2.1 T2V 1.3B as the teacher model, and \emph{image-to-video} using Wan~2.1 I2V 14B as the teacher model.
Unless otherwise specified, all non-autoregressive experiments use 16 H200 GPUs with a total batch size of 16 and AdamW optimizer.
For Wan~2.1 I2V 14B, we train our model for 800 iterations.
For Wan~2.1 T2V 1.3B, we train for 2400 iterations.
All non-autoregressive models use the same learning rate of $1\times10^{-6}$.
For the results in Tab.~\ref{tab:vbench_i2v}, Tab.~\ref{tab:quant_compare_open_models}, and the non-autoregressive ablations, we train DMD-4, DMD-8, and SC-DMD under the same setting for fair comparison.
Following DMD~\cite{yin2024one}, TMD~\cite{nie2026transition}, and CausVid~\cite{yin2025slow}, backward simulation is disabled in these non-autoregressive experiments.

\noindent\textbf{Autoregressive training.}
For autoregressive experiments, we follow the training protocols of prior real-time video generation works~\cite{huang2025self,zhu2026causal,yang2025longlive}.
We use Wan~2.1 T2V 14B as the teacher model and adopt AdamW for both the generator and the critic.
The generator learning rate is set to $2\times10^{-6}$, and the critic learning rate is set to $4\times10^{-7}$.
Backward simulation is enabled by default in autoregressive training.
As described in Sec.~\ref{sec:cache_aware}, mixed-step rollout samples $K\in\{2,4,8\}$ with probabilities $\{0.2,0.4,0.4\}$, respectively.
Standard DMD updates are applied for all sampled step counts, while the SC loss is activated only when $K=8$.
The reference alignment loss is applied to low-step rollouts and is enabled only after an initial warm-up stage for training stability; unless otherwise specified, we activate it after the model has trained for a fixed number of iterations (e.g., 600 iterations in Causal Forcing).








\section{More visualization results}

\label{sec:supp_more_visualization}

\begin{figure*}[t]
    \centering
    \includegraphics[width=1.0\linewidth]{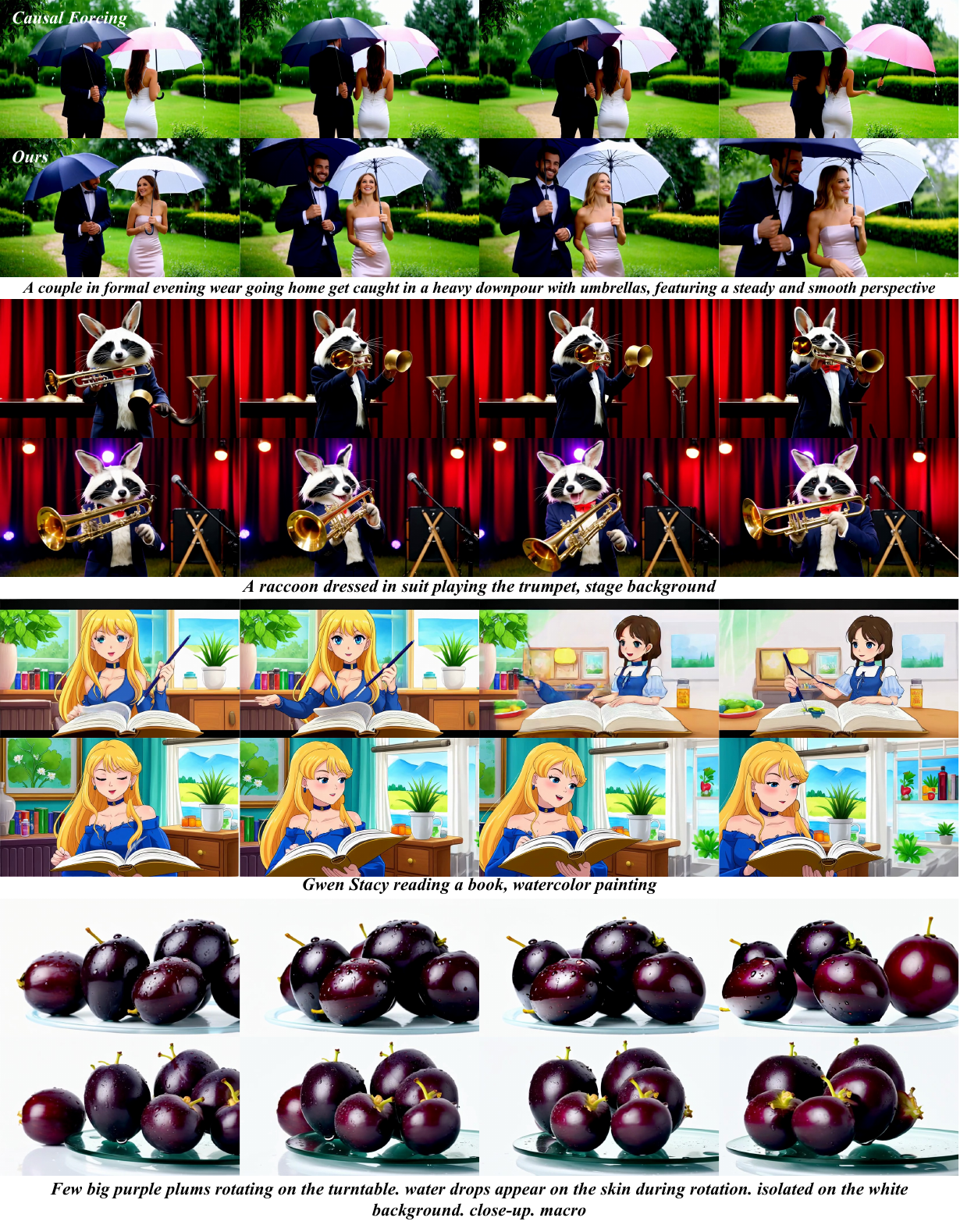}
    \caption{
    More qualitative comparisons between the Causal Forcing~\cite{zhu2026causal} baseline and our method.
    Our method shows consistent advantages in both visual quality and semantic consistency.
    Compared with the baseline, our results better preserve subject identity, object geometry, and scene composition across frames, while also producing smoother motion progression.
    The reading-girl example highlights reduced semantic/identity drift; the trombone and grape examples show improved structural stability and visual fidelity; and the umbrella example demonstrates more coherent subject interaction and temporal evolution.
    }
    \label{fig:supp_more_vis}
\end{figure*}

We provide additional qualitative comparisons between the Causal Forcing~\cite{zhu2026causal} baseline and our method in Fig.~\ref{fig:supp_more_vis}. Consistent with the quantitative improvements reported in the main paper, our method shows clearer advantages in both \emph{quality} and \emph{semantic consistency}.

First, our method better preserves subject identity and scene semantics over time. This is most clearly illustrated in \textbf{Case 3} (the reading-girl example), where the baseline exhibits obvious semantic/identity drift: the character's appearance changes substantially across frames, and the sequence even drifts toward a different character style. In contrast, our method maintains the same character identity, clothing, and scene context throughout the sequence. A similar trend can also be observed in \textbf{Case 1} (the umbrella example), where the subject pair and their interaction remain more stable under our model.

Besides, our method yields better visual quality and structural consistency across frames. In \textbf{Case 2} (the trombone example), the baseline shows stronger fluctuation in instrument geometry, body pose, and stage composition, whereas our results preserve cleaner object structure and more stable subject-object relations. In \textbf{Case 4} (the grape example), our method better maintains the spatial layout, object count, and overall composition, leading to more visually coherent and realistic videos.

What's more, our method produces smoother motion progression. In both \textbf{Case 1} and \textbf{Case 2}, the baseline often appears as a sequence of loosely related frames, while our model generates motion that evolves more naturally over time. Taken together, these examples are consistent with our claim that the proposed cache-aware training improves autoregressive rollout stability, leading to better frame quality and stronger cross-frame semantic consistency.

\end{document}